\documentclass[journal]{templates/IEEEtran}

\usepackage{soul,color}
\usepackage[pdftex]{graphicx}
\usepackage{url}
\usepackage{amssymb}
\usepackage{amsmath}
\usepackage{siunitx}
\usepackage{textcomp}
\usepackage{makecell}
\usepackage{multirow}

\usepackage{etoolbox}
\makeatletter
\patchcmd{\@makecaption}
  {\scshape}
  {}
  {}
  {}
\makeatother

\begin{document}

\title{GRIm-RePR: Prioritising Generating Important Features for Pseudo-Rehearsal}

\author{Craig~Atkinson,
        Brendan~McCane,
        Lech~Szymanski,
        and~Anthony~Robins
\thanks{Corresponding author: Craig Atkinson (atkcr398@student.otago.ac.nz)}
\thanks{Department of Computer Science, University of Otago, Dunedin, New Zealand}
}

\maketitle

\begin{abstract}
Pseudo-rehearsal allows neural networks to learn a sequence of tasks without forgetting how to perform in earlier tasks. Preventing forgetting is achieved by introducing a generative network which can produce data from previously seen tasks so that it can be rehearsed along side learning the new task. This has been found to be effective in both supervised and reinforcement learning. Our current work aims to further prevent forgetting by encouraging the generator to accurately generate features important for task retention. More specifically, the generator is improved by introducing a second discriminator into the Generative Adversarial Network which learns to classify between real and fake items from the intermediate activation patterns that they produce when fed through a continual learning agent. Using Atari 2600 games, we experimentally find that improving the generator can considerably reduce catastrophic forgetting compared to the standard pseudo-rehearsal methods used in deep reinforcement learning. Furthermore, we propose normalising the Q-values taught to the long-term system as we observe this substantially reduces catastrophic forgetting by minimising the interference between tasks' reward functions.
\end{abstract}


\IEEEpeerreviewmaketitle

\section{Introduction}

\IEEEPARstart{N}{eural} networks are capable of learning an abundance of complex tasks. Since the development of Deep Q-Networks (DQNs)~\cite{mnih2015human} they have also been able to learn relatively complex reinforcement learning tasks, including an abundance of Atari 2600 games. However, one limitation to these neural networks is that they suffer from Catastrophic Forgetting (CF)~\cite{mccloskey1989catastrophic}. This is where the neural network has the tendency to forget previously learnt knowledge while learning new information. This problem has been given increasing attention by researchers in recent years in a number of domains including image classification~\cite{kemker2017fearnet,shin2017continual,atkinson2018pseudorecursal} and reinforcement learning~\cite{rusu2016progressive,kirkpatrick2017overcoming,schwarz2018progress,kaplanis2018continual,caselles2018continual,atkinson2018pseudo}. One of the leading solutions to this problem is pseudo-rehearsal~\cite{robins1995catastrophic}. This is where a neural network learns new information while rehearsing randomly generated examples from previous tasks. Recently, this solution has been very successful when coupled with powerful generative models~\cite{shin2017continual,atkinson2018pseudorecursal,atkinson2018pseudo}. However, these generative models are currently used to generate examples without any knowledge of which features in the examples are relevant to reproduce accurately for retaining knowledge of previous tasks.

In this paper we improve the generative model used in psuedo-rehearsal by providing it information pertaining to which features it should focus on more accurately reproducing. Our results focus on using reinforcement learning to play Atari 2600 games because, due to the nature of these games, their images contain features such as sprites that provide varying amounts of useful information for playing the game.

\subsection{DQNs}
A DQN~\cite{mnih2015human} is a neural network which uses reinforcement learning to maximise its accumulative reward in a given task. More specifically, it learns to predict the expected discounted reward it will receive for taking each of its possible actions in the current state by minimising its Deep Q-Learning loss function:
\begin{equation}
\label{dqn-loss}
L = \mathbb{E}_{(s_t,a_t,r_t,s_{t+1})\sim U(D)}\Bigg[\bigg(y_t-Q(s_t,a_t;\theta_t)\bigg)^2\Bigg],
\end{equation} \begin{equation}
    y_t= 
\begin{cases}
    r_t,& \text{if terminal at } t + 1\\
    r_t + {\gamma}\max\limits_{a_{t+1}}Q(s_{t+1},a_{t+1};\theta_t^-),              & \text{otherwise}
\end{cases}
\end{equation} where the two $Q$ functions are modelled by separate deep neural networks referred to as the predictor and the target network. The predictor's parameters $\theta_t$ are updated continuously by stochastic gradient descent and the target's parameters $\theta_t^-$ are infrequently updated with the values of $\theta_t$. $(s_t,a_t,r_t,s_{t+1})\sim U(D)$ is the state, action, reward and next state for a given time step $t$ drawn uniformly from a large record of previous experiences, known as an experience replay.

\subsection{RePR}

The Reinforcement-Pseudo-Rehearsal (RePR) model~\cite{atkinson2018pseudo} extended pseudo-rehearsal methods to reinforcement learning while also utilising a generative model and a dual memory system similar to~\cite{kamra2017deep}. This allowed RePR to learn a short sequence of Atari 2600 video games without significantly forgetting how to act in previously learnt tasks. In this algorithm, memory is split into two systems; short-term memory (STM) and long-term memory (LTM). The STM system is responsible for learning the current task using deep reinforcement learning (Deep Q-Learning), while the LTM system is responsible for retaining previously learnt tasks while being taught the new task by the STM system.

The new task is taught through distillation~\cite{hinton2015distilling}, where states from the new environment (short sequence of recently seen frames/images) are inputed to the STM system to attain the desired output which the LTM system is taught to reproduce for each state. Retention of the previous tasks are achieved through pseudo-rehearsal (also know as Generative Replay~\cite{shin2017continual}), where a generative model is used to produce states representative of the previous tasks (pseudo-states) and their desired output is calculated by passing them through the previous LTM system. These input and output pairings (pseudo-items) can then be rehearsed alongside the new task. The specific loss function for training the LTM system is:
\begin{equation}
\label{repr-loss}
L_{LTM} =  \frac 1 N \sum\limits_{j = 1}^{N}\alpha {L_D}_j + (1-\alpha) {L_{PR}}_j,
\end{equation}
\begin{equation}
\label{distill-loss}
{L_D}_j = \sum\limits_{a}^{A}(Q(s_j,a;\theta_i) - Q(s_j,a;\theta_i^+))^2,
\end{equation}
\begin{equation}
\label{pr-loss}
{L_{PR}}_j =\sum\limits_{a}^{A}(Q(\widetilde{s_j},a;\theta_i) - Q(\widetilde{s_j},a;\theta_{i-1}))^2,
\end{equation} where a state $s_j$ is drawn from the experience replay of the new task and $A$ is the set of possible actions in a task. $\theta_i$ is the current weights of the long-term DQN on the new task, $\theta_i^+$ is the weights of the short-term DQN after learning the new task and $\theta_{i-1}$ is the weights of the long-term DQN after learning the previous task. Pseudo-states $\widetilde{s_j}$ are generated so that they are representative of previously learnt games. $N$ is the mini-batch size and $\alpha$ is a scaling factor weighting the importance of learning the new task compared to retaining the previous tasks via pseudo-rehearsal ($0 \leq \alpha \leq 1$).

In RePR a Generative Adversarial Network (GAN)~\cite{goodfellow2014generative} is used for generating states representative of previous games. A GAN learns to generate items representative of the examples it has been trained on by competing two networks against each other; the discriminator network learns to tell the difference between real and fake items, whereas the generative network learns to generate items which fool the discriminator into thinking that they are real. The loss functions for updating the discriminator ($L_{disc}$) and generator ($L_{gen}$) are:
\newcommand\norm[1]{\left\lVert#1\right\rVert}
\begin{equation}
\begin{aligned}
L_{disc} = D(\widetilde{x}; \phi) - D(x; \phi) + \lambda(\norm{\nabla_{\hat{x}} D(\hat{x}; \phi)}_2-1)^2 \\ + \epsilon_{drift} D(x; \phi)^2 + \epsilon_{drift} D(\widetilde{x}; \phi)^2,
\end{aligned}
\end{equation} 
\begin{equation}
\begin{aligned}
L_{gen} = -D(\widetilde{x}; \phi),
\end{aligned}
\end{equation} where $D$ is the discriminator network with the parameters $\phi$ and $G$ is the generator network with the parameters $\varphi$. $x$ is an input item which is either drawn from the current task's experience replay or the GAN of the previous long-term system. $\widetilde{x}$ is an item produced by the current generative model ($\widetilde{x} = G(z; \varphi)$) and $\hat{x} = \epsilon x + (1 - \epsilon)\widetilde{x}$. $\epsilon$ is a random number $\epsilon \sim U(0,1)$, $z$ is an array of latent variables $z = U(-1, 1)$, $\lambda = 10$ and $\epsilon_{drift} = 1e^{-6}$. The discriminator and generator network's weights are updated on alternating steps using their corresponding loss function.

\section{The GRIm-RePR Model}
Generating Representations using Importance for Reinforcement-Pseudo-Rehearsal (GRIm-RePR) improves the generator used in RePR so that it more effectively represents the previous tasks. The GAN used in RePR learns to generate states which represent the previous tasks without being provided with information about what features in the states are important for retaining knowledge of the previous games. This information could be beneficial to the GAN because it could prioritise which features it should try to more accurately generate. When the GAN's capacity for effectively learning new tasks has been exceeded, it might also provide information to the GAN about which less useful features it should forget how to generate to make room for the new task.

In RePR, the long-term system's DQN is taught how to act in all previously learnt games. This is the knowledge that we wish to retain through rehearsing generated states that effectively represent this knowledge. Therefore, sharing what features are important to this DQN with the generator will improve the effectiveness of the pseudo-states it generates. Inspired by the work of~\cite{ledig2017photo} we use the activations from an intermediate layer in the long-term system's DQN to deliver information on what features in the states are important. We chose the activations from an early layer in the neural network as early layers learn to find many useful features in input images which are combined by later layers, eventually resulting in the network's output~\cite{zeiler2014visualizing,yosinski2015understanding}. These layers will only learn features which are useful for computing the desired output in each of the tasks and therefore these activations can provide the generator information pertaining to which features are important to retain previous tasks\footnote{During early stages of testing we also tried algorithms for determining which pixels in images are more important to the long-term system's DQN than others, however we found that just using the early activation values was more effective and generally faster.}.

The question remains, how can this information be fed to the generator? Feeding this information to a generator that uses a reconstruction loss function, such as the Variational Auto-Encoder~\cite{kingma2013auto}, can be achieved trivially by adding a simple regulariser. This regulariser has the network minimise the difference between the early activation values for real and reconstructed states fed through the long-term system's DQN. Unlike GANs, we found Variational Auto-Encoders struggled to generate states from a range of games, regardless of whether this regulariser was added. Therefore, our GRIm-RePR model feeds this information to a GAN instead. This is achieved by introducing a second discriminator to the GAN which learns to classify between real and fake states from the early activation values they produce when passing them through the first two layers of the long-term system's DQN. The generator is then updated so that it produces states which fool both of the discriminators. Fooling the second discriminator is given a much higher weighting in the loss function than fooling the first discriminator, however removing the first discriminator severely impacted the generators ability to produce realistic states. The specific loss functions for the GAN used in GRIm-RePR were:
\begin{equation}
\begin{aligned}
L_{disc_2} = D(\widetilde{a}; \Phi) - D(a; \Phi) + \lambda(\norm{\nabla_{\hat{a}} D(\hat{a}; \Phi)}_2-1)^2 \\ + \epsilon_{drift} D(a; \Phi)^2 + \epsilon_{drift} D(\widetilde{a}; \Phi)^2,
\end{aligned}
\end{equation} 
\begin{equation}
\begin{aligned}
L_{gen} = -D(\widetilde{x}; \phi) -\beta D(\widetilde{a}; \Phi),
\end{aligned}
\end{equation} where the second discriminator has the parameters $\Phi$. $a=A(x; \theta)$ and $\widetilde{a}=A(G(z; \varphi); \theta)$ were $A$ returns the activations from the second layer of the long-term system's DQN using the parameters $\theta$. $\hat{a} = \epsilon a + (1 - \epsilon)\widetilde{a}$ and $\beta=1000$ in all our experiments. The discriminator and generator networks' weights were updated on alternating steps.

Furthermore, we suggest that the environments' differing reward functions interfere with each other causing continual learning to be difficult. To overcome this interference we standard normalise the Q-values from the short-term system while they are being taught to the long-term system. The mean and standard deviation used in normalisation are approximated by passing $1,000$ batches of states from the short-term system's experience replay through the short-term DQN before training the long-term system. Normalising the Q-values of the short-term system rather than the reward function of the environment is beneficial because the rewards received from the environment differ as the policy improves and thus the mean and standard deviation of the rewards received changes over time, making reinforcement learning more difficult.

\section{Related Work}
Providing additional information to a GAN to improve its generations is not an entirely new concept. For example, the conditional GAN~\cite{mirza2014conditional} achieved this by providing both the generator and the discriminator with a class label. This label told the generator which class of images it should produce an example from and this label was provided to the discriminator alongside the fake images. This allowed the discriminator to learn to tell the difference between real and fake images with prior knowledge of which class the image is suppose to belong to. Similar to our work, improving the discriminator improved the images produced by the generator.

A similar idea has also been used in continual learning~\cite{rios2019closed}, where an Auxiliary Conditional GAN was used to provide class information to the generator, while also incorporating the classification model inside of the GAN. This was more specifically achieved by having the discriminator network output $k+1$ values, where $k$ was the number of classes being classified and the additional output represented whether the network believed a generated item was real or fake. The GAN in the GRIm-RePR model does not share weights with the classifier which is advantageous because the training of the discriminator cannot interfere with the training of the classifier. Finally, the GAN used in GRIm-RePR does not require the information provided to be from a classification network and therefore it can be easily used in the reinforcement learning domain, including continuous action spaces.

In super-resolution, a low resolution image is passed through a neural network increasing its resolution. GANs have been specifically effective in this domain, where the input to the generator is replaced by a low resolution image so that the GAN learns to reconstruct a high resolution version of the image. In this task, the GAN is trying to reconstruct a specific image and therefore it has access to the real high resolution image. Therefore, in SRGAN~\cite{ledig2017photo} additional information was provided to the GAN during training by constraining the reconstructed images to produce similar features/activations than the real super-resolution images when passed through the VGG~\cite{simonyan2014very} network (pre-trained on ImageNet). Minimising the euclidean distance between these activations greatly improved the quality of the reconstructed images. A similar idea has also been applied to super-resolution of videos~\cite{lucas2019generative}, however in both of these examples the generator's desired output is known so that reconstruction loss can be used, which is not the case when training GANs for pseudo-rehearsal.

Similar methods have also been applied to Variational Auto-Encoders~\cite{kingma2013auto}. For example,~\cite{hou2017deep} trained their network to generate faces from CelebA. Their generator could produce random faces representative of the training data by passing a small number of random values as input to the generator. In this work, authors also used activations from VGG for real and reconstructed images as a constraint to improve the quality of their generations. However, this method can still not be applied directly to GANs because the generators desired output for each training example is known in Variational Auto-Encoders (as it is taught to reconstruct the images) which is not the case for GANs.

\section{Method}
We employ the same testing conditions as used in RePR~\cite{atkinson2018pseudo} where a dual learning model is taught to play Atari 2600 games in the order Pong and then Boxing. Both the DQNs (STM and LTM) are identical to~\cite{atkinson2018pseudo}, taking a sequence of four consecutive frames from the game as input and outputting the expected discounted reward from each of the possible 18 actions in the game.

As in RePR, the games are learnt by the short-term system's DQN for 20 million observable frames and then taught to the long-term system's DQN for a further 20m observable frames. When the Q-values are normalised, the long-term system's DQN is taught the first task with only $L_D$ as the loss function, whereas if the Q-values are not normalised, the first long-term system's DQN is initialised with the weights of the short-term system's DQN after learning the first task. The GAN was taught from $250,000$ sequences drawn from the previous GAN (representative of previous tasks), alongside real sequences from the current task. The GAN trains on each game for $200,000$ steps, alternating between optimising the discriminators and the generator.

The final weights of DQNs are those that performed the best while training over $250,000$ observable frames, where performance is measured by the best average reward for STM and the lowest error for LTM. Furthermore, the networks were evaluated every 1m observable frames, where the network played 30 episodes for every task, terminating an episode when all lives were lost. Actions were selected from the network with an $\epsilon$-greedy policy ($\epsilon=0.05$). The final network results are also reported using this evaluation procedure, where average rewards and standard deviations are calculated over these 30 episodes. Each condition is trained using the same set of three seeds, where all results are averaged across these three seeds.

\section{Results and Discussion}\label{results}

\begin{figure}[ht]
\vskip 0.2in
\begin{center}
\centerline{\includegraphics[trim={0cm 0cm 0cm 0cm}, clip, width=\columnwidth]{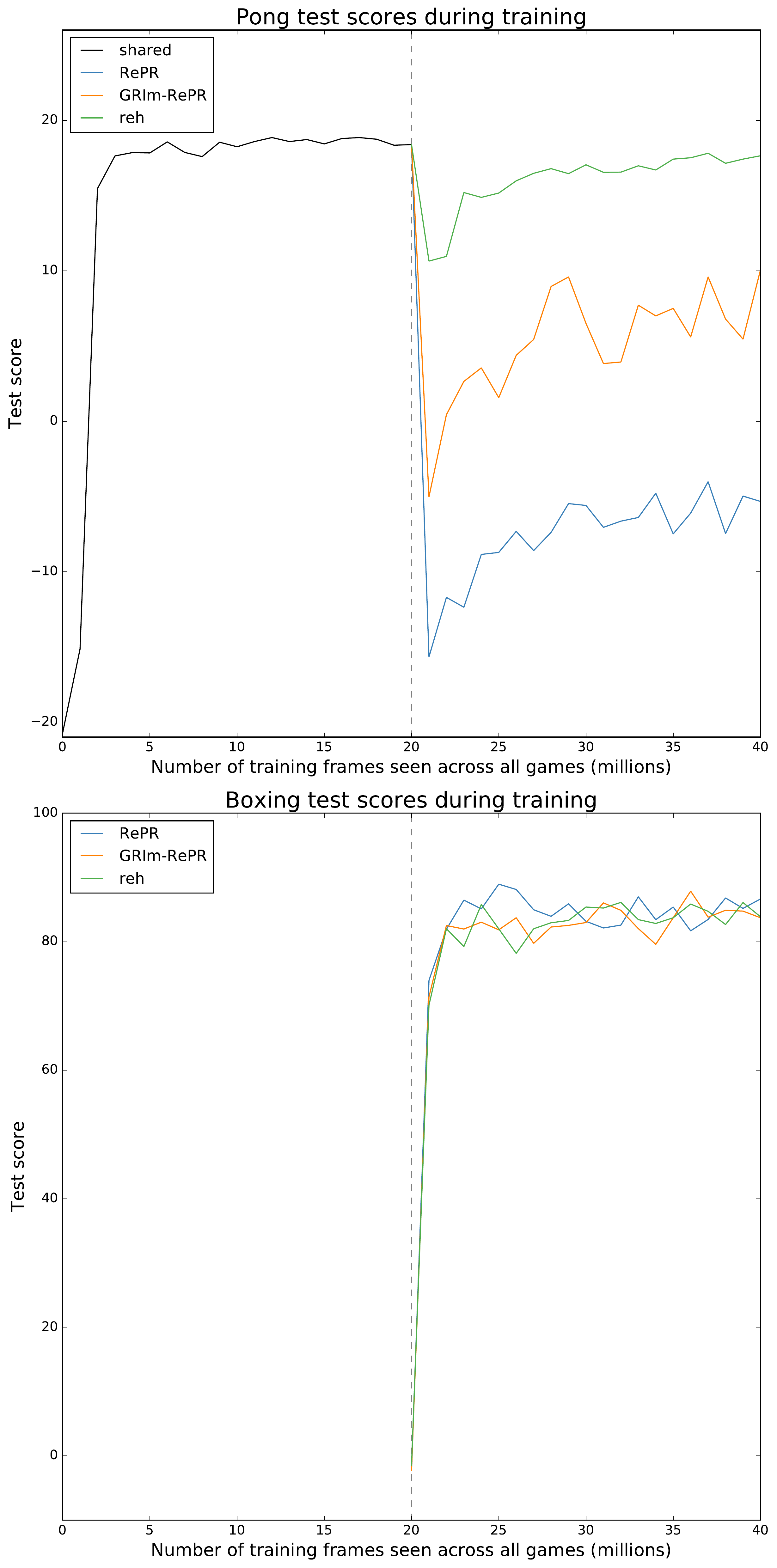}}
\caption{GRIm-RePR model compared to RePR and standard rehearsal while iteratively learning Pong and Boxing, without normalisation. Scores are recorded while training the long-term system. Task switches occur at the dashed lines, in the order Pong and then Boxing.}
\label{results-main-seed-grim}
\end{center}
\vskip -0.2in
\end{figure}

Our experiments investigated whether the addition of the second discriminator improved the retention of the game Pong when learning Boxing. Our first experiment investigated the effect without normalising the Q-values and later experiments normalise these values. Figure~\ref{results-main-seed-grim} presents the results from our improved GRIm-RePR model compared to RePR and rehearsing real items ($reh$), without normalisation. The results demonstrate that when Boxing starts to be learnt, Pong begins to be dramatically forgotten, however once Boxing has been learnt, the network begins to teach itself how to play Pong again from the pseudo-items. In the $GRIm\text{-}RePR$ condition, the model initially forgets Pong less severely than the $RePR$ condition and then begins to recovers its ability to play Pong again to a much higher performance. This demonstrates that injecting the extra information into the GAN has improved the quality of pseudo-items. The pseudo-items produced by the GRIm-RePR model are still shown to be to a lower standard than real items because the $GRIm\text{-}RePR$ condition still substantially forgets Pong compared to the $reh$ condition.

We hypothesise that the games' differing reward functions interfere with one another resulting in Pong being almost immediately forgetten. The average Q-value for Pong is approximately $2$, whereas for Boxing it is approximately $18$. This results in the loss function weighting the learning of Boxing roughly $9$ times more important than remembering Pong, explaining why Pong is so quickly forgotten. Although in this simple two task scenario, this issue could alternatively be overcome by changing the $\alpha$ parameter weighting the relative importance of distillation loss and pseudo-rehearsal loss, when extended to task sequences larger than two this is not possible and thus normalisation is necessary.

\begin{figure}[ht]
\vskip 0.2in
\begin{center}
\centerline{\includegraphics[trim={0cm 0cm 0cm 0cm}, clip, width=\columnwidth]{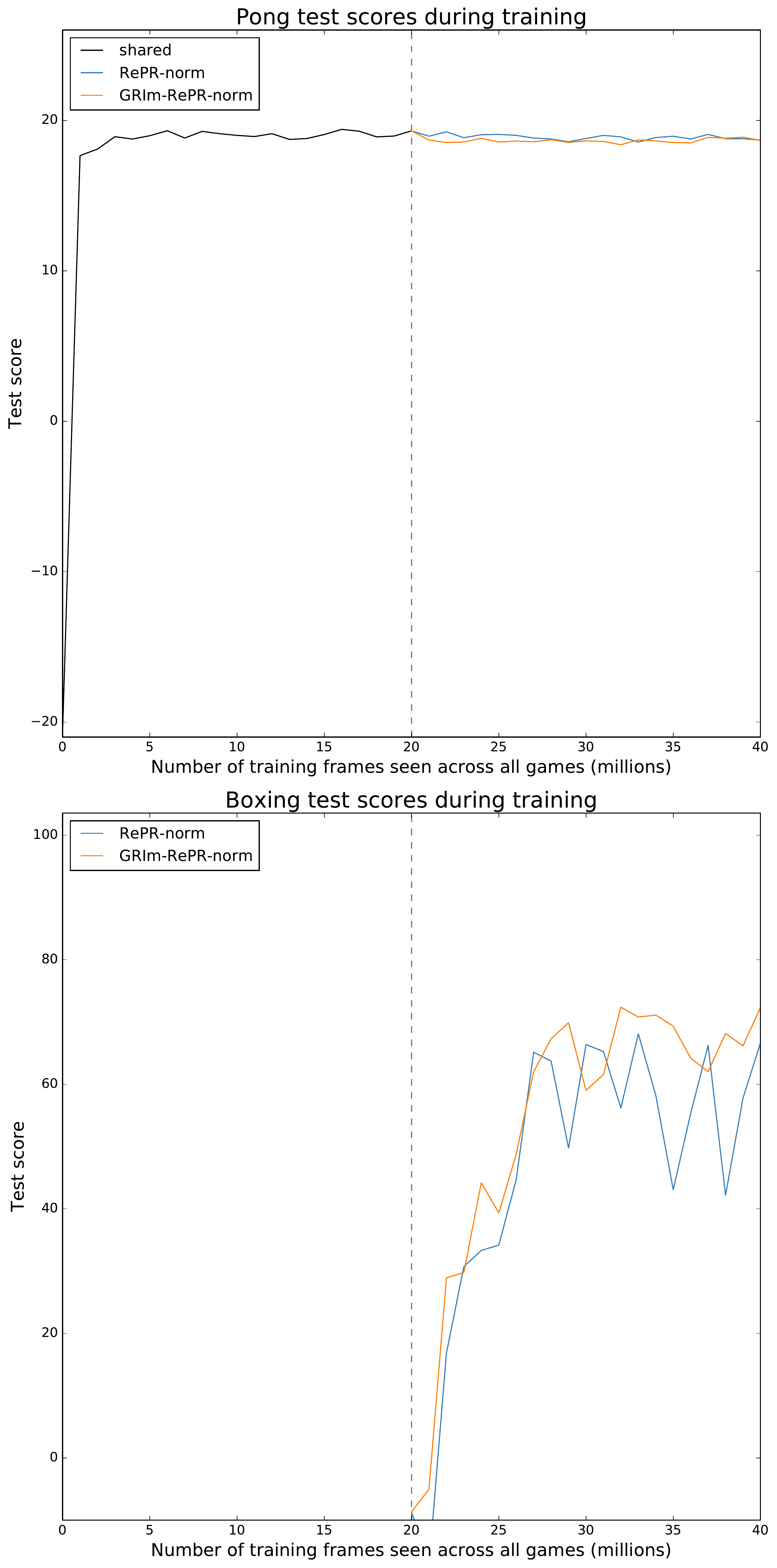}}
\caption{GRIm-RePR model compared to RePR while iteratively learning Pong and Boxing, with normalisation. Scores are recorded while training the long-term system. Task switches occur at the dashed lines, in the order Pong and then Boxing.}
\label{results-main-seed-grim-norm}
\end{center}
\vskip -0.2in
\end{figure}


Figure~\ref{results-main-seed-grim-norm} presents the results from our improved GRIm-RePR model compared to RePR, with normalisation. When normalisation is used, Pong is not quickly, nor gradually forgotten in either of the conditions, demonstrating that normalising the Q-values effectively minimised the interference between these tasks' differing reward functions. However, retaining Pong slightly interferes with the network's capability to learn Boxing, although this could potentially be due to the $\alpha$ parameter weighting retention as too important. We hypothesise that the pseudo-states produced by the GRIm-RePR model are more beneficial for pseudo-rehearsal than the pseudo-states produced by RePR, however this difference is not observable unless the long-term DQN is being challenged to remember previous tasks while learning the new task, or when the network needs to relearn previous tasks like when normalisation is not used.


Our final experiment investigates this hypothesis by using the pseudo-states produced by the generative model in both GRIm-RePR and RePR to train a randomly initialised neural network to play Pong. Firstly, the short-term system is taught to play Pong. Then the GAN is trained to generate states representative of Pong. In the GRIm-RePR conditions, the GAN uses either the same short-term system DQN to provide intermediate activations for training the second discriminator ($GRIm\text{-}RePR\text{-}norm\text{-}scratch\text{-}match$), or a different initialisation of an identical network trained to play Pong under the same conditions ($GRIm\text{-}RePR\text{-}norm\text{-}scratch\text{-}mismatch$). Pseudo-states are then generated from the model and their desired outputs are labelled by the short-term system's DQN. These pseudo-items are then exclusively used to teach a newly initialised DQN to play Pong. Simply put, the difference between the $GRIm\text{-}RePR\text{-}norm\text{-}scratch\text{-}match$ and $GRIm\text{-}RePR\text{-}norm\text{-}scratch\text{-}mismatch$ conditions is whether the DQN used for training the second discriminator and generator either matches, or does not match the DQN used for the labelling the pseudo-states used in teaching.

\begin{figure}[ht]
\vskip 0.2in
\begin{center}
\centerline{\includegraphics[trim={0cm 0cm 0cm 0cm}, clip, width=\columnwidth]{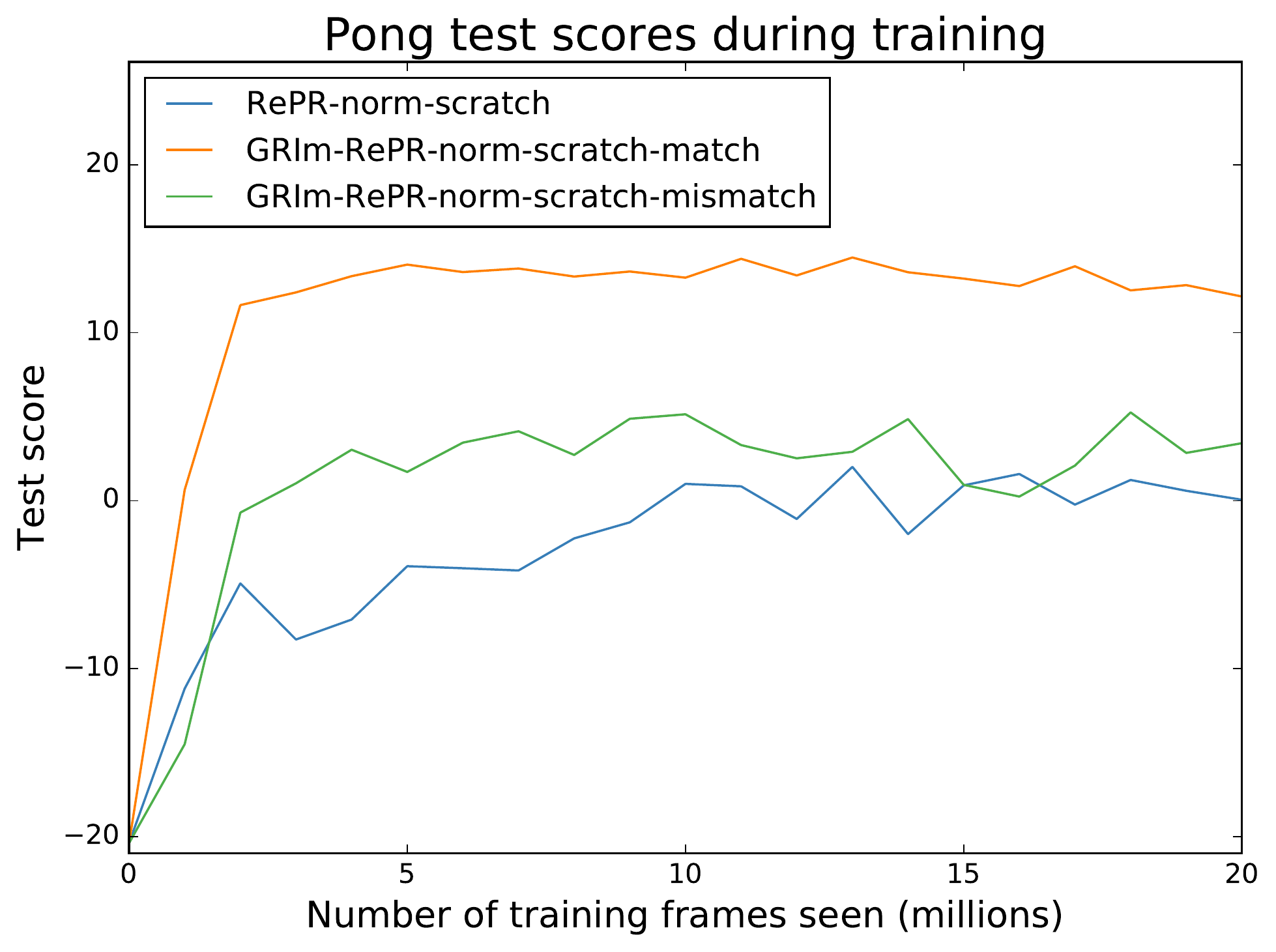}}
\caption{Results for learning Pong from scratch using input items produced by either the GRIm-RePR or RePR model and normalised outputs labelled by a teacher that has already learnt to play Pong. The match and mismatch conditions differ by whether the GRIm-RePR model has been trained using information injected from the same instance of the teacher network (used for labelling), or a separately trained one.}
\label{results-grim-one-task-match}
\end{center}
\vskip -0.2in
\end{figure}

Interestingly, the results demonstrated in Figure~\ref{results-grim-one-task-match} shows that GRIm-RePR only improves the information content of pseudo-states when the DQN used to train the GAN matches the exact DQN used to label the desired output for these states. When the DQN's do not match, GRIm-RePR performs very similarly to RePR by the end of training, potentially being quicker at learning all it can about the task from the pseudo-items. This result suggests that different instances of the DQN learn to play Pong in slightly different ways such that states specialised to one DQN, prioritises producing a different set of features that are less important to other instances of the DQN trained under the same conditions. Although, it is disappointing that the quality of the generations do not substantially improve regardless, this is still an acceptable result because it can be assumed that the same DQN will be accessible for both training the most recent GAN and labelling its pseudo-states. 

Together, our findings convey that it is beneficial to incorporate a second discriminator into the GAN to supply the generator with knowledge of which features are important to retain previously learnt tasks. However, results suggest that this improvement is only necessary when the long-term system's DQN is particularly challenged by learning the sequence of tasks. Future work should confirm whether improving the GAN results in better retention over longer and more varied task sequences. Future work might also investigate whether it is helpful to introduce a new discriminator for every intermediate layer in the agent's network or whether this knowledge can be combined or calculated in a different way.

\section{Conclusion}

In conclusion, our GRIm-RePR model improves upon the generator used in RePR by injecting information into the GAN about which features are more important to generate for retaining past tasks. This is achieved by the addition of a second discriminator, which competes against the generator. Our results demonstrate that this information improves the generated pseudo-items and thus retention, so long as the same DQN is used to inject information into the GAN as to label generated states. Furthermore, we suggest tasks' Q-values should be normalised while teaching the long-term system so to reduce the interference between the tasks' differing reward functions.

\section*{Acknowledgment}
We gratefully acknowledge the support of NVIDIA Corporation with the donation of the TITAN X GPU used for this research.

\ifCLASSOPTIONcaptionsoff
  \newpage
\fi

\bibliographystyle{templates/IEEEtran}
\bibliography{templates/IEEEabrv,mybibfile}

\begin{IEEEbiography}[{\includegraphics[width=1in,height=1.25in,clip,keepaspectratio]{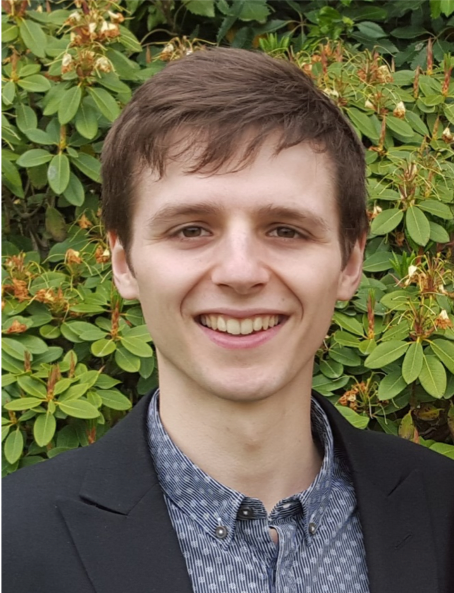}}]{Craig Atkinson}
received his B.Sc. (Hons.) from the University of Otago, Dunedin, New Zealand, in 2017. Currently, he is studying for a doctorate in Computer Science. His research interests include deep reinforcement learning and continual learning.
\end{IEEEbiography}

\begin{IEEEbiography}[{\includegraphics[width=1in,height=1.25in,clip,keepaspectratio]{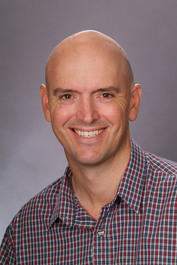}}]{Brendan McCane}
received the B.Sc. (Hons.) and Ph.D. degrees from the James Cook University of North Queensland, Townsville City, Australia, in 1991 and 1996, respectively. He joined the Computer Science Department, University of Otago, Otago, New Zealand, in 1997. He served as the Head of the Department from 2007 to 2012. His current research interests include computer vision, pattern recognition, machine learning, and medical and biological imaging. He also enjoys reading, swimming, fishing and long walks on the beach with his dogs.
\end{IEEEbiography}

\begin{IEEEbiography}[{\includegraphics[width=1in,height=1.25in,clip,keepaspectratio]{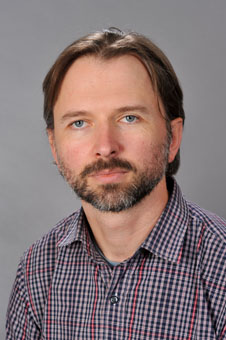}}]{Lech Szymanski}
received the B.A.Sc. (Hons.) degree in computer engineering and the M.A.Sc. degree in electrical engineering from the University of Ottawa, Ottawa, ON, Canada, in 2001 and 2005, respectively, and the Ph.D. degree in computer science from the University of Otago, Otago, New Zealand, in 2012.   He is currently a Lecturer at the Computer Science Department at the University of Otago. His research interests include machine learning, artificial neural networks, and deep architectures.
\end{IEEEbiography}

\begin{IEEEbiography}[{\includegraphics[width=1in,height=1.25in,clip,keepaspectratio]{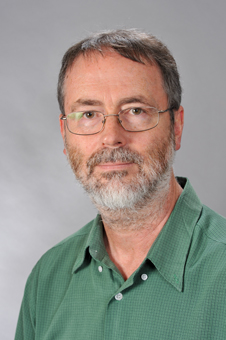}}]{Anthony Robins}
completed his doctorate in cognitive science at the University of Sussex (UK) in 1989.  He is currently a Professor of Computer Science at the University of Otago, New Zealand.  His research interests include artificial neural networks, computational models of memory, and computer science education.
\end{IEEEbiography}

\end{document}